\pdfoutput=1
\documentclass[10pt,twocolumn,letterpaper]{article}

\usepackage[accsupp]{axessibility}

\usepackage{iccv}
\usepackage{times}
\usepackage{epsfig}
\usepackage{graphicx}
\usepackage{amsmath}
\usepackage{amssymb}

\newcommand\blfootnote[1]{
  \begingroup
  \renewcommand\thefootnote{}\footnote{#1}
  \addtocounter{footnote}{-1}
  \endgroup
}

\usepackage{authblk}
\makeatletter
\renewcommand\AB@affilsepx{\hspace{3mm} \protect\Affilfont}
\makeatother

\usepackage[colorlinks,bookmarks=false]{hyperref}
\usepackage[hyphenbreaks]{breakurl}

\iccvfinalcopy 

\def\iccvPaperID{1}

\ificcvfinal\fi

\begin{document}

%%%%%%%%% TITLE
\title{The Multi-Modal Video Reasoning and Analyzing Competition}

%%%%%%%%% Author section start

\author[1,2]{Haoran Peng}
\author[1]{He Huang}
\author[1]{Li Xu}
\author[1]{Tianjiao Li}
\author[1]{Jun Liu}
\author[2]{Hossein Rahmani}
\author[3]{Qiuhong Ke}
\author[4]{Zhicheng Guo}
\author[5]{Cong Wu}
\author[5]{Rongchang Li}
\author[6]{Mang Ye}
\author[4]{Jiahao Wang}
\author[6]{Jiaxu Zhang}
\author[6]{Yuanzhong Liu}
\author[7]{Tao He}
\author[8]{Fuwei Zhang}
\author[9]{Xianbin Liu}
\author[8]{Tao Lin}

\affil[1]{Singapore University of Technology and Design} 
\affil[2]{Lancaster University}
\affil[3]{University of Melbourne}
\affil[4]{Xidian University}
\affil[5]{Jiangnan University}
\affil[6]{Wuhan University}
\affil[7]{Tsinghua University \hspace{15mm} }
\affil[8]{Sun Yat-sen University}
\affil[9]{BOE Technology Group Co., Ltd}

%%%%%%%%% Author section end

\maketitle

\ificcvfinal\thispagestyle{empty}\fi

%%%%%%%%% ABSTRACT
\begin{abstract}
In this paper, we introduce the Multi-Modal Video Reasoning and Analyzing Competition (MMVRAC) workshop in conjunction with ICCV 2021. This competition is composed of four different tracks, namely, video question answering, skeleton-based action recognition, fisheye video-based action recognition, and person re-identification, which are based on two datasets: SUTD-TrafficQA and UAV-Human. We summarize the top performing methods submitted by the participants in this competition, and show their results achieved in the competition.
\blfootnote{
Haoran Peng (\nolinkurl{phr@mails.ccnu.edu.cn}), 
He Huang (\nolinkurl{he\_huang@mymail.sutd.edu.sg}), 
Li Xu (\nolinkurl{li\_xu@mymail.sutd.edu.sg}), 
Tianjiao Li (\nolinkurl{tianjiao\_li@mymail.sutd.edu.sg}), 
Jun Liu (\nolinkurl{jun\_liu@sutd.edu.sg}), 
Hossein Rahmani (\nolinkurl{h.rahmani@lancaster.ac.uk}), 
and Qiuhong Ke (\nolinkurl{qiuhong.ke@unimelb.edu.au}) 
are the MMVRAC 2021 challenge organizers. The MMVRAC 2021 website: \url{https://sutdcv.github.io/multi-modal-video-reasoning}.}
\end{abstract}

%%%%%%%%% BODY TEXT
\section{Introduction}

Visual relational reasoning is a crucial element of human reasoning yet a challenging task for computer vision algorithms. While most of the existing works have focused on reasoning from still images, understanding visual relationships in videos has received limited attention. However, videos enable us to reason about more comprehensive spatio-temporal relationships.

To promote the research development in the multi-modal video reasoning and analyzing area, we organize the first ICCV workshop competition on Multi-Modal Video Reasoning and Analyzing. In this competition, we have four tracks, namely, Track-1: video question answering, Track-2: skeleton-based action recognition, Track-3: fisheye video-based action recognition, and Track-4: person re-identification. The goal of Track-1 is to answer questions about the video content by performing spatio-temporal and logical reasoning over the video content. Track-2 and Track-3 respectively aim to recognize human behaviours from skeleton data and RGB videos captured by an ultra wide-angle fisheye camera. The goal of Track-4 is to perform the person re-identification task under a flying UAV. 

The SUTD-TrafficQA dataset \cite{Xu_2021_CVPR}, which is a traffic event-based video reasoning dataset with two modalities (RGB videos and texts), is used to evaluate the methods for Track-1.
The UAV-Human dataset \cite{Li_2021_CVPR} containing multiple modalities, such as skeletons, RGB and fisheye videos, is used to evaluate the methods for Track-2, Track-3 and Track-4. 

The remainder of this paper is organized as follows: Section 2 describes the two datasets used for the challenge. Section 3 introduces four different tracks and Section 4 summarizes the results of the challenge. Section 5 introduces the top performing methods for each track. Finally, we conclude the paper in Section 6.

%-------------------------------------------------------------------------
\section{Datasets}

\subsection{SUTD-TrafficQA Dataset}
SUTD-TrafficQA is a traffic event-based video reasoning dataset. It contains 10,080 in-the-wild videos and 62,535 question answering (QA) pairs about event understanding of complex traffic scenarios. The dataset provides six challenging traffic related reasoning tasks, including 1) basic understanding, 2) event forecasting, 3) reverse reasoning, 4) counterfactual inference, 5) introspection and 6) attribution. More details of this dataset can be found in \cite{Xu_2021_CVPR}.

\subsection{UAV-Human Dataset}
UAV-Human is a large-scale human behaviour understanding dataset collected by a flying UAV. This dataset contains 22,476$\times$3 video sequences for human action recognition, 22,476 images for human pose estimation, 41,290 images and 1,144 identities for human re-identification (Re-ID), and 22,263 images for human attribute recognition. It includes different data modalities, including RGB videos, depth videos, IR sequences, skeleton data, fisheye videos and night-vision videos, to enable human behavior analysis under different conditions. More details of the UAV-Human dataset can be found in \cite{Li_2021_CVPR}.

%-------------------------------------------------------------------------

\section{Competition}

We have hosted the first Multi-Modal Video Reasoning and Analyzing Competition (MMVRAC) in conjunction with ICCV 2021 to encourage the development of the state-of-the-art video reasoning and understanding methods. Specifically, we provided four different tracks for visual reasoning from videos.

%-------------------------------------------------------------------------
\subsection{Track 1: Video Question Answering}

Video question answering is an important research topic among vision-and-language tasks. It focuses on answering the questions about the video content. To well understand the logical reasons of the events in videos, the models need to analyze the video content in spatio-temporal way. This problem is a hot topic in the computer vision area, which however is quite challenging, and the models' performance in video question answering still need to be further improved.

%-------------------------------------------------------------------------
\subsection{Track 2: Skeleton-based Action Recognition}

Skeleton-based action recognition is a task about recognizing human behaviors through skeletal data. This task attracts many attentions since skeleton data is very concise and can also represent human behaviors. UAV viewpoints are very useful in real world application such as city surveillance and catastrophe rescue. In these situations, UAV is the common equipment, and thus, understanding human action through UAV viewpoints becomes a very important task.

%-------------------------------------------------------------------------
\subsection{Track 3: Fisheye Video-based Action Recognition}

Fisheye video-based action recognition is the task of understanding human actions from the videos that are captured by an ultra wide-angle camera. The fisheye camera can provide broad views, which is useful for many real-world UAV application scenarios. However, there have been very few work about recognizing human actions from fisheye videos. Thus this task is still largely open.

%-------------------------------------------------------------------------
\subsection{Track 4: Person Re-Identification}

Person re-identification is the task of identifying whether there is a specific pedestrian in the captured images or video sequences. It is a very challenging yet very important task that has a wide range of applications in intelligent video surveillance, intelligent security, and other fields. Person re-identification from a flying UAV is even more challenging but also useful in real world applications.

%-------------------------------------------------------------------------

%------------------------------------------------------------------------
\section{Results of Competition}

There were 53 submissions in Track 1, 74 submissions in Track 2, 46 submissions in Track 3, and 68 submissions in Track 4. The results achieved by the top 3 ranked teams of each track are shown in Tables 1, 2, 3, and 4. %\ref{table1}, \ref{table2}, \ref{table3}, and \ref{table1}.
%------------------------------------------------------------------------

\begin{table}
\begin{center}
\caption{The results of the Top-3 teams for Track 1: video question answering. Note that the evaluation set used in this competition is different from the testing set in \cite{Xu_2021_CVPR}.}

\begin{tabular}{|l|c|c|}
\hline
Rank & Team & Accuracy\\
\hline\hline
1 & IPIU\_VQA & 48.3\% \\
2 & Go For It & 36.7\% \\
3 & mote & 33.7\% \\
\hline
\end{tabular}
\end{center}

\end{table}

\begin{table}
\begin{center}
\caption{
The results of the Top-3 teams for Track 2: skeleton-based action recognition. Note that the evaluation set used in this competition is different from the testing set in \cite{Li_2021_CVPR}.}
\begin{tabular}{|l|c|c|}
\hline
Rank & Team & Accuracy\\
\hline\hline
1 & A Rowing Boat & 51.5\% \\
2 & 322Win & 49.9\% \\
3 & CRIPAC & 49.3\% \\
\hline
\end{tabular}
\end{center}
\end{table}

\begin{table}
\begin{center}

\caption{The results of the Top-3 teams for Track 3: fisheye video-based action recognition. Note that the evaluation set used in this competition is different from the testing set in \cite{Li_2021_CVPR}.}
\begin{tabular}{|l|c|c|}
\hline
Rank & Team & Accuracy\\
\hline\hline
1 & A Rowing Boat & 45.4\% \\
2 & t322 & 37.0\% \\
3 & BOE\_AIOT\_AIBD & 35.9\% \\
\hline
\end{tabular}
\end{center}
\end{table}

\begin{table}
\begin{center}

\caption{The results of the Top-3 teams for Track 4: Person Re-Identification.
}
\begin{tabular}{|l|c|c|}
\hline
Rank & Team & Accuracy\\
\hline\hline
1 & MARS\_WHU & 79.1\% \\
2 & MIG & 74.5\% \\
3 & ISEE-ACW & 72.2\% \\
\hline
\end{tabular}
\end{center}
\end{table}

\section{\texorpdfstring{Competition Methods}{}}

This section introduces the methods of the top performing teams for each track. The descriptions of the methods are provided by the corresponding teams.

%------------------------------------------------------------------------

\subsection{Methods of Video Question Answering}

%------------------------------------------------------------------------

\subsubsection{Rank 1: Team IPIU\_VQA}

The team, IPIU\_VQA, with 5 members (Yuhan Wang, Xinyu Liu, Ting Su, Zhicheng Guo and Licheng Jiao) submitted the solution which is based on ClipBERT \cite{lei2021more}.%, an end-to-end training network. 

Firstly, the resolution of each input video was changed to $448 \times 448$. For the vision encoder, the team employed ResNet-50, which is initialized with weights from grid-feat \cite{ren2016faster}. Specifically, the team used the first five convolutional blocks of ResNet-50, and added a convolution layer to reduce the depth of the output feature. A $2 \times 2$ max-pooling layer was used for spatial down-sampling. Avg-pooling was used as a temporal fusion layer and the resulting feature map was flattened into an embedding sequence for representing the clip, which contains 144 pixels \cite{lei2021more}. 

The team used a trainable word embedding layer as their language encoder to encode language tokens. There were also trainable position embeddings to encode the position information of the tokens. Then the team used different types of embeddings for both clip and text embeddings \cite{2018BERT} to indicate their source type. Thus, these two sequences were concatenated as the input of a 12-layer transformer \cite{2017Attention} for cross-modal fusion. 

According to the concept of ``less is more'' \cite{lei2021more}, the team sparsely sampled the clips at the training phase. The team used three types of sampling clips: 4 clips (2 frames per clip), 8 clips (1 frame per clip), and 8 clips (2 frames per clip) for training. Its effect is higher and more reliable than a dense sampling of the entire content of the video. 

The team utilized three pre-training strategies. On the one hand, the team adopted large-scale image-text datasets (COCO Captions \cite{2015Microsoft} and Visual Genome Captions \cite{2016Visual}) to perform cross-modal pre-training, and the ClipBERT weights can be obtained directly from \textit{ https://github.com/jayleicn/ClipBERT}. 
On the other hand, the team used the above initialized ClipBERT weights to train on TGIF-QA action/transition \cite{jang2017tgifqa} and VQA v2 \cite{Goyal_2017_CVPR}, and its weights were also used for the training of video question answering tasks. The impact of different weights initialization strategies showed to be beneficial by experiments. Then the team fine-tuned their model from these three types of pre-trained weights for downstream video-text tasks.
%------------------------------------------------------------------------

\subsubsection{Rank 2: Team Go For It}

The team, Go For It, with 4 members (Jiahao Wang, Wang Hao, Yifei Chen and Fang Liu) divided the dataset into training and validation sets in the ratio of 8:2, and then ran through the baseline HCRN network \cite{le2020hierarchical} at the beginning of the competition. For extracting video features, the team used the pre-training models, namely ResNet50, ResNet101, and ResNet152, for video feature extraction, respectively. The team cut the video into 8 segments, and extracted 16 frames per segment as feature images with the size of $(b,8,16,2048)$. For extracting text features, the team used Glove to encode the word vectors, and fed the text features and video features together into the HCRN network for question answering prediction. For the determination of the number of categories, the team used the length of the concatenated set of all question answers in the training set as the size of the number of categories in the last layer of the HCRN (the total number of categories was 501). 

In the middle of the competition, the team extracted the features from  HCRN before the input linear layer into traditional machine learning (e.g., SVM, random forest, and xgboost) for training, and finally fused the results with other results.

Later in the competition, the team merged the previously divided training set and the validation set to form a new training set and re-trained the models. Finally, the team counted the frequencies of all the answers to the questions and selected the unpredicted answers among all the fused results according to their frequencies, and the highest frequency answer was used as the final result to improve the final prediction result.

%------------------------------------------------------------------------

\subsubsection{Rank 3: Team mote}
The team, mote, with 3 members (Fuwei Zhang, Duo Chen and Mingjie Zhou) used the pre-trained ResNet to extract the appearance features of the video, and then used the pre-trained ResNeXt (resnext-101-kinetics) to extract the motion features of the video, and processed the video into appearance and motion channels, respectively. Their model consists of four parts: appearance graph attention module, motion graph attention module, local-to-global attention module, and global-to-local module. Among them, the team took local-to-global and global-to-local modules as their core method, and appearance graph attention and motion graph attention as their supplementary modules.

The specific process of the local-to-global attention module is as follows. The team first used questions to pay attention to motion and appearance features, respectively. Due to the strong correlation between adjacent frames in the same video, the team used 1×1 convolution to separately pay attention to the fusion features. Since the same visual feature of the same video would be consistent at different times, the team used multi-head self-attention to local attention feature map doing global attention calculation. The specific process of the global-to-local attention module is as: because the same video and the same visual entity change with the time dimension. The team first adopted a multi-head self-attention mechanism to pay global attention to the fusion features of the two channels. In the same calculation, the team used 1x1 convolution to do the local attention calculation on the global attention feature map. Finally, the team concatenated the features of the four modules to get their candidate features.
%------------------------------------------------------------------------

\subsection{Methods of Skeleton-based Action Recognition}
%------------------------------------------------------------------------

\subsubsection{Rank 1: Team A Rowing Boat}

The team named A Rowing Boat has 7 members (Cong Wu, Zhongwei Shen, Rongchang Li, Tianyang Xu, Xiao-Jun Wu, Josef Kittler and Jiwen Lu). 

Recent studies have shown that due to the non-Euclidean structure of skeleton data \cite{yan2018spatial}, Graph Neural Networks provides intrinsic superiority in modeling spatio-temporal skeleton information. 
Based on this, the team used their recently proposed method Graph2Net \cite{wu2021graph2net}, which is an efficient and effective graph-based method, as the basic model. 
Considering the characteristics of the UAV-Human dataset \cite{Li_2021_CVPR}, such as large variations of perspectives and high similarities of some classes, the team proposed an effective solution to handle the problem of skeleton-based action recognition. 

Their solution includes three major stages: data processing, training, and ensemble. 
At the data processing stage, a common observation is that different feature representations specify certain characteristics in distinguishing different actions. 
Thus, the team used the following feature representations, \textit{i.e.}, joint, bone, and angular \cite{qin2021leveraging}, to perform discriminative analysis and modeling of skeleton sequences from multiple perspectives. 
At the training stage, to endow prior information into the model, the team performed pre-training on a large-scale skeleton dataset. 
According to the distribution of commonly used skeleton datasets, the team chose kinetics-skeleton \cite{kay2017kinetics} as the pre-training dataset, and performed appropriate pre-processing to guarantee its consistency with the data format in this competition. 
To explore the identification of difficult classes and improve the robustness of the model, focal loss \cite{lin2017focal} and label smooth \cite{szegedy2016rethinking} were also used. 
Besides, the fusion of classification scores from different models can often boost the final performance, which is in line with the common sense. 
Therefore, in the ensemble stage, the team considered the complementary differences among multiple models. 
First of all, the models obtained by the aforementioned different feature representations and training strategies exhibit different concerns. 
For instance, though focal loss can strengthen the inter-class discrimination of specific classes, it also delivers a negative impact on the modeling of some samples. 
Furthermore, feature modeling for different temporal scales can often obtain multi-granularity feature representations.
Hence, it is reasonable to integrate the classification information at different granularities in the method.
The team also utilized the successful elements of some state-of-the-art methods, including Shift-GCN \cite{cheng2020skeleton} and MS-G3D \cite{liu2020disentangling}, in their method. 

%------------------------------------------------------------------------

\subsubsection{Rank 2: Team 322Win}
The team, 322Win, with 6 members (Jiaxu Zhang, Jinlu Zhang, Zhisheng Huang, Yuanzhong Liu and Zhigang Tu) designed a data pre-processing method for noisy skeleton data and adopted a multi-stream fusion strategy. The team used MS-G3D \cite{liu2020disentangling} as their baseline model. Specifically, their multi-stream model consists of the following 7 streams: (1) A 2D-joint stream, which takes 2D joint coordinates of the human body as the input data. (2) A 2D-bone stream, which takes 2D bone vectors of the human body as the input data. The bone vectors can be obtained by calculating the first-order spatial difference of the joint coordinates. (3) A 2D-velocity stream, which takes the 2D velocity vectors of the human joints as the input data. The velocity vectors can be obtained by calculating the first-order temporal difference of the joint coordinates. (4) A 3D-joint stream, which takes 3D joint coordinates of the human body as the input data. The 3D joint coordinates data is reconstructed from the 2D data through the VideoPose3D \cite{Pavllo_2019_CVPR} model. This 3D reconstruction process can effectively reduce the noise of the data and provide effective 3D information. (5) A 3D-bone stream, which 
uses 3D bone vectors of the human body as the input data. (6) A 3D-velocity stream, which uses 3D velocity vectors of human joints as the input data. (7) A pre-trained 2D-joint stream, which uses the Kinetics-Skeleton dataset to pre-train the MS-G3D model.

%------------------------------------------------------------------------
\subsection{Methods of Fisheye Video-based Action Recognition}

\subsubsection{Rank 1: Team A Rowing Boat}
The team named A Rowing Boat has 7 members (Rongchang Li, Cong Wu, Zhongwei Shen, Tianyang Xu, Xiao-Jun Wu, Josef Kittler and Jiwen Lu). 

Compared to common benchmarks, the fisheye camera and UAV platform bring more challenges, such as distortions, camera shaking, resolution variations, etc. Besides, the UAV-Human dataset contains action classes with different granularity levels, and these actions occur in a variety of scenarios, which poses a higher requirement of the algorithm’s ability in understanding videos. To overcome these challenges, the team has made efforts in both data and models. Their solution can be divided into three phases: data processing, feature extraction, and prediction ensemble. At the data processing phase, the team sought to alleviate the inherent video quality problems of the fisheye camera and UAV platform. The team utilized center cropping and data augmentation tricks to mitigate the effects of camera shake, resolution variations, various illumination, and image distortion. At the feature extraction stage, the team attempted to comprehend video sequences from different perceptual perspectives. Specifically, the team used two video-level sampling strategies: one is a sparse sampling method where 24 frames are uniformly sampled from each video, and the other is a dense sampling method where 16 × (5 continuous frames) are uniformly sampled for each video. For the former, the team proposed an innovative graph model (GM) to extract global features according to the unstructured distribution of sparse temporal points. For the latter, the team used TDN \cite{wang2021tdn} to extract dense features. Since the dataset contains various actions, objects, and scenes, etc., the team needs to use larger datasets to provide effective prior knowledge to cover these patterns. Based on the observation that pre-training with different datasets will produce different types of knowledge, the team selected motion-focused Something-Something V2 \cite{mahdisoltani2018effectiveness} and scene-focused Kinetics-400 \cite{carreira2017quo} to train the GM and TDN models. During transfer training, the team employed label smooth \cite{szegedy2016rethinking} to calculate the cross-entropy loss and replaced the final average pooling layer with Gempooling \cite{radenovic2018fine} layer to improve generalization. Following the above thinking line, the team got a dual network architecture containing four models that were respectively transferred from two datasets. In the prediction ensemble section, the team first employed multiple views ($n$ clips × $m$ crops) to improve the inference performance of a single model, and then the team integrated the predictions of the four models mentioned above. It is worth mentioning that the team attempted more modality (optical flow) and 3D model (slow-fast \cite{Feichtenhofer_2019_ICCV}) and expected it could gain some patterns they had neglected. Finally, combining these two complementary solutions resulted in a small improvement ($0.8\%$). 
%------------------------------------------------------------------------

\subsubsection{Rank 2: Team t322}
The team named t322 has 6 members (Yuanzhong Liu, Ke Li, Beiming Chang, Jinlu Zhang, Jiaxu Zhang and Zhigang Tu). 

A multi-model-based two-stream framework was adopted for fisheye video-based action recognition. Their proposed framework consists of two parts: fisheye image rectification and two-stream neural network. For fisheye image rectification, a Progressively Complementary Network \cite{Yang_2021_CVPR} was utilized to correct the deformation in the fisheye video frames. For two-stream neural network, origin and rectified RGB frames were used for spatial stream, and TVL1 optical flow \cite{ipol.2013.26} extracted from origin fisheye video frames were used for temporal stream. SlowFast \cite{Feichtenhofer_2019_ICCV} and SlowOnly \cite{Feichtenhofer_2019_ICCV} models were pre-trained on the Kinetics400 \cite{carreira2017quo} dataset, TANet \cite{DBLP:journals/corr/abs-2005-06803} was pre-trained on the Something V1 dataset \cite{Goyal_2017_ICCV}. The scores of the four models were averaged to obtain the final predictions. 
%------------------------------------------------------------------------

\subsubsection{Rank 3: Team BOE\_AIOT\_AIBD}
The team, BOE\_AIOT\_AIBD, with 3 members (Xianbin Liu, Zeyu Shangguan and Zhanfu An) observed that the actions often appear in the central part of the videos in the dataset. Therefore, the team first used the original video and left the central part to relieve the severe barrel distortion of the data and chased off the irrelevant information accordingly, which make the training process more effective and thus convergent rapidly. 
The team chose the CNN baseline and tried the 3D convolution algorithm with 2 pre-trained models: SlowFast and X3D. In addition, the team proposed a special strategy to sampling the videos at various intervals, that is, sampling the videos at different frequencies so that the team did not miss any information. Then the team fed these sampling results to the network separately to extract action features. After the team got the results of both models mentioned above, the team fused them by calculating the weighted average as the final prediction. The team further plugged in the channel attention model so that their network would concentrate more on the body actions in the video. 
Furthermore, the team applied the feature fusion between feature maps and thus efficiently divined the location information in the shallow layers, as well as the semantic information in the deep layer. Their proposal also enriched the semantic in higher-level feature maps, which aligns with the fact that action recognition depends more on higher-level feature maps. 
The fused model proposed by the team performed much better than the single model: the test results are 0.332 for SlowFast, 0.338 for X3D, and 0.359 for this fused model.
The team implemented this result on 2 v100 with epochs of 50, batch size of 32, and learning rate of 0.5 and 0.005 for SlowFast and X3D, respectively. No extra supplementary datasets were used while training. 
%------------------------------------------------------------------------

%------------------------------------------------------------------------
\subsection{Methods of Person Re-Identification}

\subsubsection{Rank 1: Team A MARS\_WHU}
The team named MARS\_WHU has 5 members (Mang Ye, Shuoyi Chen, Tongxin Wang, He Li and Bo Du).

The dual-stream network contains a transformer-based network and a CNN-based network. The transformer has a strong ability to focus on the human features under low lighting conditions where CNN suffers from noise contained in low lighting images. The CNN can learn pixel-wise features where the transformer cannot learn the information inside each patch. Both networks were trained independently. The distance matrices of the two networks are combined at the test phase.

In the transformer stream, the Vision Transformer pre-trained on ImageNet was used as backbone initialization. Their network contains two branches. The first branch utilizes the global feature, and the second branch adopts shuffled PCB features. The loss function is defined as:
\begin{equation}
    \centering
    Loss = \mathcal{L}_{id}(f_g) + \mathcal{L}_{tri}(f_g) + 
    \frac{1}{N}\sum_{i = 1}^{N}({\mathcal{L}_{id}(f_i) + \mathcal{L}_{tri}(f_i)})
\end{equation}

In the CNN stream, ResNet-101-IBN-A was used as the backbone and the parameters were pre-trained on ImageNet.
To combine the respective advantages of CNN and transformer, the team performed a weighted fusion of the distance matrices obtained by the CNN and transformer methods to obtain the final result. More specifically, the fusion weight of Transformer based distance matrix is $0.9$.
UAV-Human contains around 500 training samples with label noise. The noisy samples were manually picked out from the training set (about $1\%$ performance gain). The images were resized to 384x128 and augmented with padding 10 pixels, random cropping, random horizontal flipping, color jitter, and random erasing.

The widely-used Re-rank method was also applied. Because most identities in the test set only contain two target images and each identity have two queries, the Re-rank parameters were set to $k_1 = 4, k_2 = 4, \lambda = 0.45$. Re-ranking optimized the initial retrieval ranking list by exploiting the relationship between the gallery and query samples.

To enhance the robustness against pose or variations, the team applied the multi-shot query setting by aggregating the calculated distances of different queries from the same identity. This strategy improved the performance by fully utilizing the rich query information.

%------------------------------------------------------------------------

\subsubsection{Rank 2: Team MIG}
The team, MIG, with 2 members (Tao He and Leqi Shen) adopted the widely used open-source framework \textit{fast-reid}~ \cite{DBLP:journals/corr/abs-2006-02631}.

\noindent\textbf{Data: }
A small part of the whole training set was first divided as the validation set. After determining the hyper-parameters on it, all the training data were combined to train the final model. Long-tailed data (person ID with only one image) was also removed from the training set. As for the input size, $256 \times 128$, $384 \times 128$, or $384 \times 192$, were evaluated. Random horizontal flip, random erasing, and auto-augmentation were used as data augmentation.

\noindent\textbf{Model: }
Three kinds of backbones were used: ResNet101-ibn-a(R101-ibn)~ \cite{Pan_2018_ECCV}, ResNeXt101(X101)~ \cite{Xie_2017_CVPR}, and ResNeSt101(S101)~ \cite{DBLP:journals/corr/abs-2004-08955}. The \textit{non-local} module was also adopted in the backbone. The last pooling layer was replaced with \textit{gem pooling}. On top of the backbone, a BN layer and classification layer were added~ \cite{Luo_2019_CVPR_Workshops}. Cross-entropy loss, triplet loss with soft margin, and circle loss were simultaneously used to update the model. The team also tested the model pre-trained on ImageNet and open-source Re-ID dataset, such as MSMT17~ \cite{Wei_2018_CVPR}. Offline results on the validation set showed the ImageNet pre-trained models were better.

\noindent\textbf{Training Details: }
The team randomly sampled 4 instances per person in a mini-batch, resulting in batch size 64. Adam optimizer with 0.00035 learning rate was used. A warm-up strategy is adopted for the first 2000 iterations. The backbone was frozen in the first 1000 iterations. The learning rate was constant in the first 30 epochs and is decayed by cosine annealing scheduler in the next 30 epochs. 

\noindent\textbf{Evaluation: }
Three models with varied input size were ensembled to generate the final distance matrix on the test set: R101-ibn(384x192), S101(384x128), X101(256x128), X101(384x128). Re-ranking was also used as post-processing before model ensemble. To further boost the performance, the team also generated examples with high confidence from the test set as training data (604 IDs with 1799 images). As a result, two additional models, X101(256x128) and S101(384x128) were also added to the ensemble models.

The work by the team MIG is supported by the National Natural Science Foundation of China (Nos. U1936202).

%------------------------------------------------------------------------

\subsubsection{Rank 3: Team ISEE-ACW}
The team named ISEE-ACW has 4 members (Tao Lin, Xiao Li, Chengzhi Lin and Ancong Wu).

The team found that the competition has two key challenges: With different clothes, the same person will be regarded as different IDs; Many images have low light illumination such that it is hard to identify people. So the team proposed a local-and-global method that not only focuses on the global image but also the local part of the image.

First, the team trained a strong pre-trained model based on the code of FastReID. The training datasets include MSMT17, CUHK03, DukeMTMC, Market1501 and the training dataset of this competition. The backbone was ResNet-101. 

Next, the team fine-tuned the pre-trained model using two different approaches. One is normal training, which uses classification loss and triplet loss. For the other, the team used a head-shoulder adaptive attention network (HAA) to solve the Black Re-ID problem. The HAA would assigning a larger weight on the head-shoulder feature if the image's individual is wearing black clothing.
To get the head-shoulder bounding box, the team split roughly the top third of the images. The former focuses on the global image, and the latter focuses on the local part of the image. 

Reranking is another useful tool in this competition. The team found that automatic query expansion (AQE) has a validation map improvement of $1.2\%$.

%------------------------------------------------------------------------

\section{Conclusion}
In this paper,
we reported the first Multi-Modal Video Reasoning and Analyzing Competition (MMVRAC) in conjunction with ICCV 2021.
This competition aims to encourage the development of novel and effective approaches to improve the capability of video reasoning and understanding. There were hundreds of participants and submissions which produced many interesting and powerful solutions.
We are glad to congratulate the winner teams for achieving great results and their interesting methods. We would also like to thank all the participants for their efforts and contributions in the video reasoning and analyzing area.

\section*{Acknowledgement} 
This work is supported by the SUTD Project PIE-SGP-Al2020-02 and the TAILOR project funded by EU Horizon 2020 research and innovation programme under GA No 952215. The authors would like to thank all the volunteers and participants of this competition. The authors also thank the teams that provided the descriptions of their methods, and the members of these teams include Xinyu Liu, Ting Su, Yuhan Wang, Licheng Jiao, Zhongwei Shen, Tianyang Xu, Xiao-Jun Wu, Josef Kittler, Jiwen Lu, Shuoyi Chen, Tongxin Wang, He Li, Bo Du, Wang Hao, Yifei Chen, Fang Liu, Jinlu Zhang, Zhisheng Huang, Zhigang Tu, Ke Li, Beiming Chang, Leqi Shen, Duo Chen, Mingjie Zhou, Zeyu Shangguan, Zhanfu An, Xiao Li, Chengzhi Lin and Ancong Wu.

%------------------------------------------------------------------------

{\small
\bibliographystyle{ieee_fullname}
\bibliography{egbib}
}

\end{document}